%% file: main-010.tex
\documentclass[runningheads]{llncs}

\usepackage{graphicx}
\usepackage{comment}
\usepackage{amsmath,amssymb}
\usepackage{color}
\usepackage{url}
\usepackage{hyperref}
\usepackage{multirow}
\usepackage{tablefootnote}
\usepackage{array}
\usepackage[toc,page]{appendix}

\usepackage{stfloats}
\fnbelowfloat

\newif\ifreview
\reviewtrue
\reviewfalse

\ifreview
	\usepackage{lineno}

	\linenumbers
\fi

\begin{document}


\def\SubNumber{010}

\def\GCPRTrack{Young Researcher's Forum}

\title{Efficient and Discriminative Image Feature Extraction for Universal Image Retrieval}

\ifreview
\else
	\titlerunning{Image Feature Extraction for Universal Image Retrieval}

	\author{Morris Florek\inst{1}\orcidID{0009-0008-8425-5161} \and
    David Tschirschwitz\inst{1}\orcidID{0000-0001-5344-4172} \and
    Björn Barz\inst{2}\orcidID{0000-0003-1019-9538} \and
    Volker Rodehorst\inst{1}\orcidID{0000-0002-4815-0118}}
    \authorrunning{M. Florek et al.}
	
	\institute{Bauhaus-University Weimar, 99423 Weimar, Germany \email{morris.benedikt.florek@uni-weimar.de} \and 
    Carl Zeiss AG, 07745 Jena, Germany}
    
\fi

\maketitle              

\begin{abstract}

Current image retrieval systems often face domain specificity and generalization issues. This study aims to overcome these limitations by developing a computationally efficient training framework for a universal feature extractor that provides strong semantic image representations across various domains. To this end, we curated a multi-domain training dataset, called \textit{M4D-35k}, which allows for resource-efficient training. Additionally, we conduct an extensive evaluation and comparison of various state-of-the-art visual-semantic foundation models and margin-based metric learning loss functions regarding their suitability for efficient universal feature extraction. Despite constrained computational resources, we achieve near state-of-the-art results on the Google Universal Image Embedding Challenge, with a \(mMP@5\) of 0.721. This places our method at the second rank on the leaderboard, just 0.7 percentage points behind the best performing method. However, our model has 32\% fewer overall parameters and 289 times fewer trainable parameters. Compared to methods with similar computational requirements, we outperform the previous state of the art by 3.3 percentage points. We release our code and \textit{M4D-35k} training set annotations at \url{https://github.com/morrisfl/UniFEx}.

\keywords{Image Retrieval \and Universal Features \and Compute Efficient.}

\end{abstract}

%
%
\section{Introduction}

The prevalence of image capturing devices has led to the growth of digital image collections and the need for advanced image retrieval systems. Content-based image retrieval (CBIR) finds semantically similar images from a large database given a query image \cite{li_recent_2021}. CBIR has many applications in various fields: it speeds up medical image searches in emergencies \cite{qayyum_medical_2017}, assists e-commerce shoppers in finding similar products \cite{zhang_visual_2018}, helps locate and identify landmarks \cite{zhang_landmark_2018}, and enables law enforcement to identify individuals for safety purposes \cite{jain_face_2012}. However, current methods are often limited by their domain-specificity \cite{noh_large-scale_2017,cao_unifying_2020} and encounter difficulties with out-of-domain images and lack of generalization. Since the utilization of multiple per-domain models in a unified image retrieval system is both costly and inconvenient \cite{feng_unifying_2020}, a unified model capable of retrieving images across multiple domains is desirable.

\begin{figure}[t]
\centering
\includegraphics[width=0.85\textwidth]{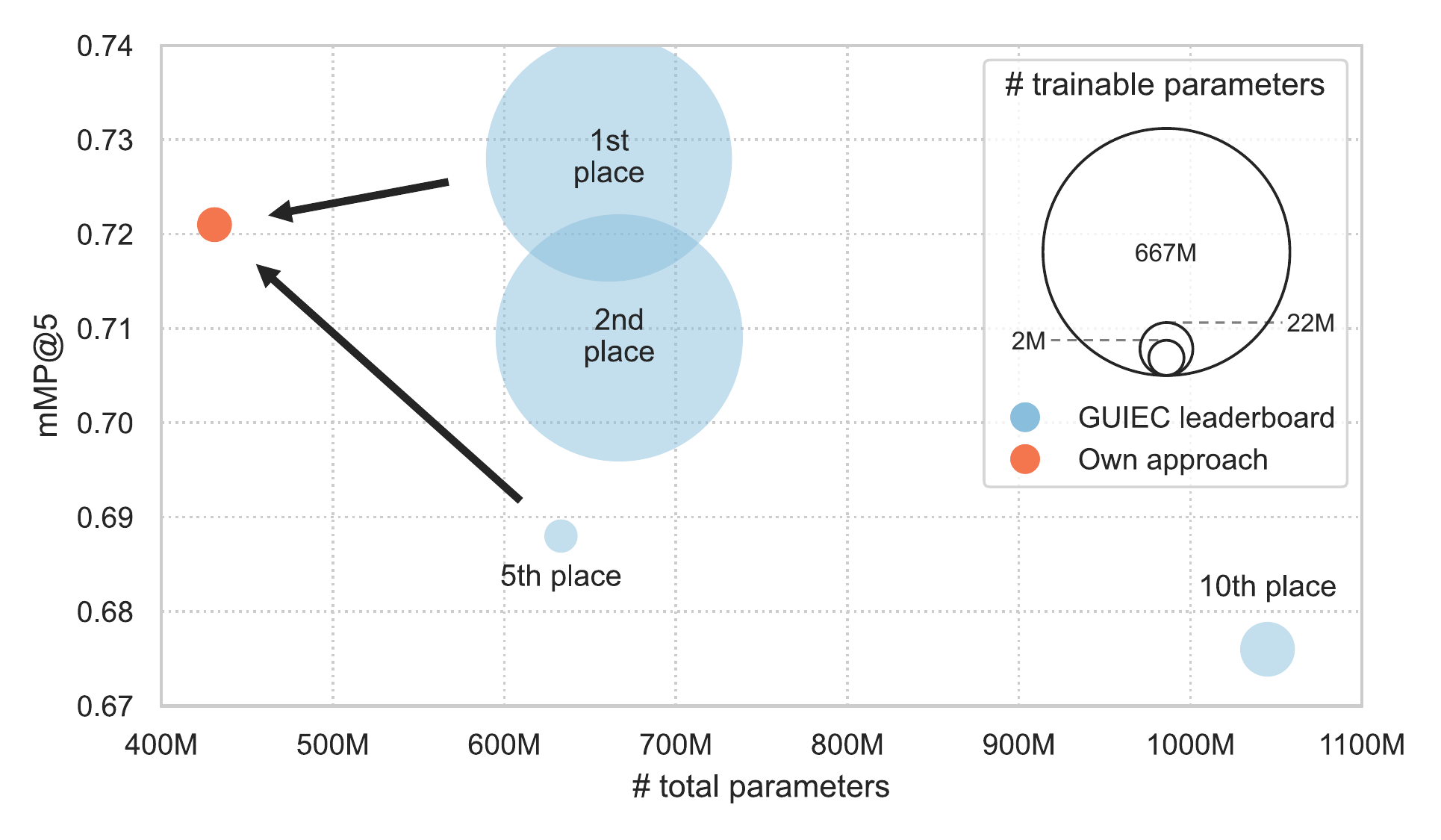}
\caption{Results on the GUIEC \cite{araujo_google_2022} test set. Comparing our approach to the GUIEC leaderboard by plotting the evaluation metric (\(mMP@5\)) over the number of total model parameters. The bubble’s area is proportional to the number of trainable model parameters.} 
\label{fig:contribution}
\end{figure}

Recognizing that the universal capabilities of retrieval systems depend on the image representation, this study delves into the realm of universal feature extraction. Therefore, the primary objective was to efficiently develop and train a universal image encoder capable of extracting discriminative image features specifically tailored for image retrieval at the instance-level. We present two distinct contributions: (1) \textit{M4D-35k}, a streamlined multi-domain training set, allowing for resource-efficient training. Unlike existing multi-domain training sets, it supports supervised learning, features instance-level class labeling, and a more balanced domain and class distribution. (2) Substitution studies on the efficacy of various visual-semantic foundation models and margin-based metric learning losses, identifying the optimal combination for universal image representation learning. This resulted in a close to State-Of-The-Art (SOTA) result on the Google Universal Image Embedding Challenge (GUIEC) \cite{araujo_google_2022}, as shown in Figure \ref{fig:contribution}, while using significantly less computational resources for training by solely fine-tuning the projection head (i.e., linear probing).

%
%

\section{Related Work}

\subsubsection{Fine-grained Multi-domain Datasets.} Fine-grained datasets have a detailed label classification structure, resulting in a large number of distinct classes and a long-tailed class distribution. At the most detailed level of classification, these labels correspond to specific objects, architectural structures, or scenes, delineating instance-level characteristics. Although many datasets are available at the fine-grained \cite{russakovsky_imagenet_2015,krause_3d_2013,bossard_food-101_2014} or instance-level \cite{bai_products-10k_2020,weyand_google_2020,liu_deepfashion_2016}, they are often limited to specific image domains. Conversely, to the best of our understanding, there remains a scarcity of fine-grained multi-domain datasets that are suitable for training a universal image encoder for retrieval purposes.

Table \ref{tab:datasets} lists existing multi-domain datasets, along with their scope and characteristics, including their level of class granularity, domain count, and dataset size. Contrary to these datasets, which incorporate fine-grained classified samples, our \textit{M4D-35k} training set comprises solely instance-level data. Unlike the smaller-sized INSTRE \cite{wang_instre_2015} and GPR1200 \cite{schall_gpr1200_2022} datasets, which were primarily designed for evaluative purposes, the focus of \textit{M4D-35k} is on the training of universal image representations. While the MRT \cite{almazan_granularity-aware_2022} dataset is partitioned into equally sized training and test sets, the training set is unlabeled. In contrast, \textit{M4D-35k} is fully labeled, enabling supervised learning methodologies.

During the editing phase of this study, UnED \cite{ypsilantis_towards_2023}, a new large-scale, multi-domain dataset, was released. UnED integrates images from publicly available datasets across eight domains, and offers distinct training, validation, and test splits. Its training set consists of 2.8M samples across 316k classes, marked by an unbalanced class distribution, with about half of the samples derived from a single data source. In contrast, \textit{M4D-35k} is tailored for resource-efficient training. It contains a curated selection of 328k images spanning 35k classes, ensuring a more balanced class distribution and diversified representation of data sources.

\begin{table}[t]
\caption{Comparison of our \textit{M4D-35k} training set with existing multi-domain datasets in terms of their scope and dataset characteristics.}
\label{tab:datasets}
\centering
\resizebox{\textwidth}{!}{%
\begin{tabular}{l|l|c|c|c|c}
\hline
\textbf{Dataset} & \textbf{Scope}          & \textbf{Granularity}     & \textbf{\# domains$ $}               & \textbf{\# classes$ $} & \textbf{\# images} \\ \hline
INSTRE \cite{wang_instre_2015}             & Evaluation                          & Fine-grained         & 3                   & 200                 & 23k                \\
GPR1200 \cite{schall_gpr1200_2022}         & Evaluation                          & Fine-grained         & 6                   & 1.2k                & 12k                \\
MRT \cite{almazan_granularity-aware_2022}  & \textbf{Training} \& evaluation $ $ & Fine-grained         & 6                   & 23k                 & 267k               \\
UnED \cite{ypsilantis_towards_2023}        & \textbf{Training} \& evaluation     & Fine-grained         & \textbf{8}          & \textbf{349k}       & \textbf{4.1M}       \\ \hline
\textit{M4D-35k} (ours) $ $                & \textbf{Training}                   & \textbf{Instance-level}    & 4                   & 35k                 & 328k  \\ \hline             
\end{tabular}
}
\end{table}

\subsubsection{Universal Image Representation.} In 2022, Kaggle hosted the GUIEC \cite{araujo_google_2022}, a competition focused on developing cutting-edge strategies and techniques for training universal image representations. These representations were intended for efficient retrieval of images across multiple domains. Participants proposed different methodologies, which were evaluated using a disclosed evaluation set. This set contained 200k index and 5k query images, covering 11 different image domains, and was split equally into a validation (public score) and test (private score) set. The modified Mean Precision at 5 (\(mMP@5\)) was employed to evaluate the performance of the submitted approaches.

Leading approaches used a pre-trained OpenCLIP \cite{ilharco_openclip_2021} foundation model as a backbone with an attached projection head to comply with the 64-dimensional embedding constraint of the challenge. These models underwent supervised training on a custom multi-domain dataset, using either ArcFace \cite{deng_arcface_2019} or Sub-Center ArcFace \cite{deng_sub-center_2020} as the loss function. The top two approaches \cite{shao_1st_2022,huang_2nd_2022} fine-tuned their models end-to-end, treating the backbone and projection head differently, either through a multi-stage approach or by using different learning rates. Notably, the teams that placed 5th \cite{ota_5th_2022} and 10th \cite{koo_10th_2022} only trained the projection head and kept the backbone frozen. The 5th place added normalized input image dimensions (width, height, and aspect ratio) to the backbone embeddings, while the 10th place fused embeddings from OpenCLIP encoders of different sizes.

Following the leading methods \cite{shao_1st_2022,huang_2nd_2022}, we constructed our image embedding model. Our approach integrates a visual-semantic foundation model as backbone, complemented by a projection head, and utilizes a margin-based metric learning loss. This study, however, ventures beyond by evaluating the efficacy of a variety of foundation models and margin-based losses in the context of universal image representation. Unlike the top two approaches \cite{shao_1st_2022,huang_2nd_2022}, we only trained the projection head (i.e., linear probing) owing to computational constraints. Therefore, we used the training settings from the 5th \cite{ota_5th_2022} and 10th \cite{koo_10th_2022} places, while outperforming these approaches and obtaining close to SOTA results with 289$\times$ less trainable parameters than the top-ranking work.

%
%

\section{Universal Image Representation}
\subsection{\textit{M4D-35k} Dataset}

\begin{figure}[!b]
  \centering
  \begin{minipage}[]{0.65\textwidth}
    \centering
    \resizebox{\linewidth}{!}{%
    \begin{tabular}{cllcc}
    \hline
    \textbf{Rank} & \textbf{Dataset}          & \textbf{Domain}        & \textbf{\# uses} & \textbf{\textit{mAP}} \\ \hline
            1             & Products-10k \cite{bai_products-10k_2020}             & Products         & 15               & 0.548     \\
            2             & GLDv2 (cleaned) \cite{weyand_google_2020}   & Landmarks              & 12               & 0.377     \\
            3             & DeepFashion \cite{liu_deepfashion_2016}            & Fashion & 6                & 0.208     \\
            4             & MET Artwork \cite{ypsilantis_met_2022}               & Artwork                & 7                & 0.194     \\
            5             & Shopee \cite{howard_shopee_2021}                    & Products         & 3                & 0.141     \\
            6             & H\&M Personalized Fashion \cite{garcia_ling_hm_2022} & Fashion & 3                & 0.073      \\
            7             & RP2k \cite{peng_rp2k_2021}                      & Products         & 4                & 0.056      \\
            8             & Stanford Online Products \cite{song_deep_2016}  & Fashion & 3                & 0.052      \\
            9             & Fashion-200k \cite{han_automatic_2017}              & Fashion & 3                & 0.052      \\
            10            & Food Recognition 2022 \cite{alcrowd_aicrowd_2022}     & Dishes                 & 4                & 0.051     \\
            11            & Stanford Cars \cite{krause_3d_2013}             & Cars                   & 3                & 0.048      \\
            12            & DeepFashion2 \cite{ge_deepfashion2_2019}              & Fashion & 2                & 0.038      \\
            13            & Food101 \cite{bossard_food-101_2014}                   & Dishes                 & 2                & 0.025      \\ \hline
    \end{tabular}%
    }
  \end{minipage}
  \hfill
  \begin{minipage}[]{0.32\textwidth}
    \includegraphics[width=\linewidth]{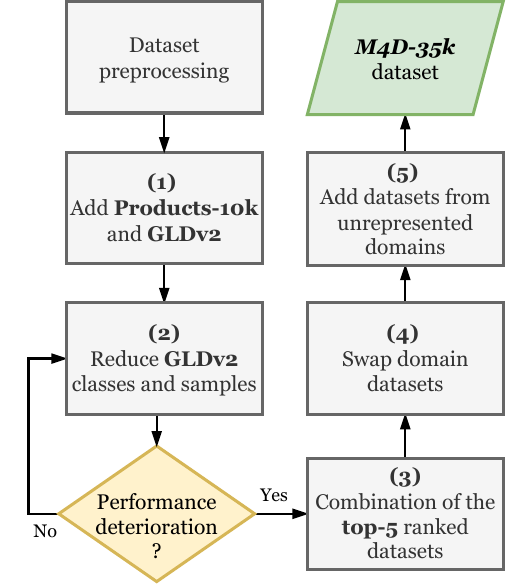}
  \end{minipage}
  \caption{The table on the left displays the datasets considered for the curated \textit{M4D-35k}. Datasets are ranked according to their frequency of use in the GUIEC \cite{araujo_google_2022}, as measured by the \(mAP\) relative to the GUIEC leaderboard rank. The curation process is shown on the right.}
  \label{fig:m4d-35k}
\end{figure}

Given the absence of a suitable pre-existing multi-domain training set at instance-level, we curated our own dataset. The aim was to curate a multi-domain training set from publicly available datasets that facilitate resource efficient training. The selection and incorporation of datasets was guided by linear probing an image embedding model on various dataset configurations and evaluating its performance on the GUIEC \cite{araujo_google_2022} validation set. We refer to this multi-domain training set as \textit{M4D-35k}.

\subsubsection{Data Sources.} The datasets under consideration, which were integral to the curation process, are shown in Figure \ref{fig:m4d-35k}. These datasets were selected based on their utilization by the top-performing teams in the GUIEC \cite{araujo_google_2022}. The selection criteria comprised dataset availability, usage frequency (minimum of two instances), and employment by at least one of the top-5 approaches. To assess the significance of each dataset, the mean Average Precision (\(mAP\)) relative to the GUIEC leaderboard rank was computed. A total of 13 datasets were identified, covering six different domains. Combining all 13 datasets would result in an extensive collection of approximately 3.36M images across 503k classes, making resource-efficient training unfeasible. Consequently, we optimized the scope of the training data by strategically minimizing its size.

\subsubsection{Preprocessing.} Prior to the curation process, an initial preprocessing of all datasets was performed. This was conducted to achieve a more balanced distribution of samples across classes. Classes containing fewer than three samples were discarded, and those exceeding 100 samples were randomly downsized to a maximum of 100 samples each. Furthermore, it became apparent during the curation process that the inclusion of instance-level datasets was beneficial. Therefore, an additional preprocessing step was performed on the Stanford Cars \cite{krause_3d_2013} dataset to refine its class granularity. A car color classification model, EfficientNet-B1 \cite{tan_efficientnet_2019}, pre-trained on the ImageNet-1K \cite{russakovsky_imagenet_2015}, was fine-tuned on the Vehicle Color Recognition \cite{panetta_artificial_2021} dataset. At inference, this model was employed to predict the colors of vehicles, leading to a finer-grained classification where each class represents a unique combination of car model and color. This contrasts with the previous classification, which was based solely on car models.

\subsubsection{Curation Process.} The dataset curation process was divided into five stages, as shown in Figure \ref{fig:m4d-35k}. At each stage, different dataset configurations were used to linearly probe the image embedding model. The model architecture is depicted in Figure \ref{fig:model-arch}. We utilized the OpenCLIP ViT-H/14 \cite{ilharco_openclip_2021}, pre-trained on the Laion-2B \cite{schuhmann_laion-5b_2022}, as the backbone, along with the ArcFace \cite{deng_arcface_2019} loss function. The embedding model underwent linear probing for a total of 2.56M viewed samples. The performance was evaluated on the GUIEC \cite{araujo_google_2022} validation set, with the highest \(mMP@5\) being the primary metric used to guide our decision-making process. Further details regarding the evaluation results can be found in \nameref{sec:data-cur}.

Owing to their frequent use in the GUIEC \cite{araujo_google_2022}, the Products-10k \cite{bai_products-10k_2020} and GLDv2 \cite{weyand_google_2020} datasets were pre-selected for inclusion in the \textit{M4D-35k} training set and thus were not subjected to subsequent selection processes. Nevertheless, we attempted to downsize the GLDv2 dataset by examining the total class volume and the upper threshold for class samples. Through the analysis of diverse configurations, we ascertained an optimal arrangement comprising 10k classes, each with a maximum of 10 samples. This configuration effectively reduced the size of the initial GLDv2 dataset by an estimated 94.4\%, while maintaining performance, resulting in a more resource efficient training set.

\noindent Furthermore, the synergistic effects of different dataset combinations were analyzed, focusing on the top-5 datasets according to their ranking. With the inclusion of Products-10k \cite{bai_products-10k_2020} and GLDv2 \cite{weyand_google_2020} subset fixed, all feasible combinations with DeepFashion \cite{liu_deepfashion_2016}, MET Artwork \cite{ypsilantis_met_2022}, and Shopee \cite{howard_shopee_2021} were evaluated. The integration of DeepFashion and Shopee individually resulted in the most favorable outcomes. This led us to explore alternative datasets within the same domain to identify potential improvements. Consequently, DeepFashion was substituted by H\&M Personalized Fashion \cite{garcia_ling_hm_2022}, Fashion-200k \cite{han_automatic_2017}, and DeepFashion2 \cite{ge_deepfashion2_2019}, while Shopee was replaced by RP2k \cite{peng_rp2k_2021} and Stanford Online Products \cite{song_deep_2016}. However, these adjustments did not result in any performance improvements, leaving the configuration consisting of Products-10k, GLDv2 subset, and DeepFashion as the most effective and diverse. 

Finally, we incorporated datasets from unrepresented domains to expand the domain variety. This included the Food Recognition 2022 \cite{alcrowd_aicrowd_2022} and Food101 \cite{bossard_food-101_2014} datasets from the dishes domain, as well as Stanford Cars \cite{krause_3d_2013} from the cars domain. The integration of the dishes datasets failed to produce any discernible improvements, which may be attributed to the broader class classification granularity inherent in these datasets. However, the inclusion of Stanford Cars, especially its refined version, resulted in substantial performance gains. This highlights the significance of instance-level class characteristics in the \textit{M4D-35k} training set.

\begin{table}[t]
\caption{Final configuration of the \textit{M4D-35k} training set, with the included dataset, its domain, and its size in terms of number of classes and images.}
\label{tab:m4d-35k}
\centering
\begin{tabular}{l|l|c|c}
\hline
\multicolumn{1}{l|}{\textbf{Domain}} & \multicolumn{1}{l|}{\textbf{Dataset}} & \textbf{\# classes} & \textbf{\# images} \\ \hline
Products           & Products-10k \cite{bai_products-10k_2020}         & 9.5k                & 141.5k             \\
Landmarks          & GLDv2 \cite{weyand_google_2020} (subset)          & 10.0k               & 79.2k              \\
Fashion            & DeepFashion \cite{liu_deepfashion_2016}           & 14.3k               & 100.4k             \\
Cars               & Stanford Cars \cite{krause_3d_2013} (refined)     & 1.0k                & 7.3k               \\ \hline
Multi-Domain       & \textit{M4D-35k}                                  & 34.8k               & 328.4k     \\ \hline       
\end{tabular}
\end{table}

\subsubsection{\textit{M4D-35k}.} The \textit{M4D-35k} training set is sourced from four public available datasets—Products-10k \cite{bai_products-10k_2020}, a GLDv2 \cite{weyand_google_2020} subset, DeepFashion \cite{liu_deepfashion_2016}, and the refined Stanford Cars \cite{krause_3d_2013}—and encompasses four distinct domains. Through strategic dataset selection and the implementation of strict criteria for the total class volume and sample thresholds per class, we have successfully compressed the size of the training data. As shown in Table \ref{tab:m4d-35k}, the training set comprises 328k images distributed among 35k distinct instance-level classes. This represents a selection of less than 10\% of the initial 3.36M samples, achieved without compromising model performance, thereby facilitating a more resource-efficient training procedure.

\subsection{Image Embedding Model}

\begin{figure}[t]
\centering
\includegraphics[width=1\textwidth]{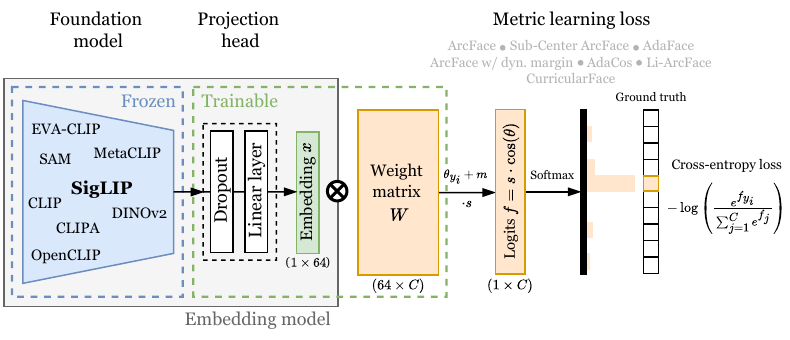}
\caption{The embedding model consists of a visual-semantic foundation model as backbone, followed by a projection head.  During training, a margin-based metric learning loss is employed, with cosine similarities \(\cos(\theta)\) derived via matrix multiplication from the  normalized embeddings \textit{x} and weights \textit{W}. An angular margin \textit{m} is added to the target angle \(\theta_{y_i}\), logits are scaled by the scaling parameter \textit{s}, and both softmax activation and cross-entropy loss are applied. The model’s trainable and non-trainable components are also detailed.} 
\label{fig:model-arch}
\end{figure}

The model's architectural concept was inspired by the best practices \cite{shao_1st_2022,huang_2nd_2022} observed in the GUIEC \cite{araujo_google_2022}, as shown in Figure \ref{fig:model-arch}. The architecture includes a pre-trained visual-semantic foundation model that serves as the backbone for extracting robust, general-purpose image embeddings. A projection head, comprising a dropout layer (dropout rate of 0.2) and a linear layer, is built on top of the backbone embeddings to compress them into a 64-dimensional space. During training, a margin-based metric learning loss is employed to enhance the discriminative power of the embeddings. In order to address computational constraints, the training process was limited to the projection head of the embedding model (i.e., linear probing), which required us to freeze the entire backbone and set us apart from the leading methods \cite{shao_1st_2022,huang_2nd_2022} of the GUIEC, which fine-tuned their entire model. During the experimental phase of this research (refer to Section \ref{sec:experiments}), a series of substitution studies were conducted to assess the effectiveness of various visual-semantic foundation models and margin-based metric learning losses in the context of universal feature learning.

\subsubsection{Foundation Model.} Foundation models are models mostly trained on diverse data through self-supervision at scale, possessing the flexibility to adapt to a wide range of downstream tasks \cite{bommasani_opportunities_2022}. Among these, image-text contrastive learning approaches, such as CLIP \cite{radford_learning_2021}, OpenCLIP \cite{ilharco_openclip_2021}, CLIPA \cite{li_inverse_2023}, EVA-CLIP \cite{sun_eva-clip_2023}, MetaCLIP \cite{xu_demystifying_2023}, or SigLIP \cite{zhai_sigmoid_2023} possess excellent zero-shot classification capabilities. Additionally, DINOv2 \cite{oquab_dinov2_2023}, a self-supervised paradigm, has demonstrated performance on par with CLIP models in linear probing scenarios. The Segment Anything Model (SAM) \cite{kirillov_segment_2023}, has achieved impressive outcomes in zero-shot segmentation tasks. These models primarily employ a Vision Transformer (ViT) \cite{dosovitskiy_image_2020} architecture as their visual component for image encoding. We considered them in this study, since they span different pre-training paradigms and are strong candidates for deriving robust and universal image embeddings.

\subsubsection{Metric Learning Loss.} Margin-based metric learning losses represent a modification of the conventional softmax loss. They include a margin penalty, which serves to enhance the discriminative capacity of the image embeddings. SOTA methods, such as ArcFace \cite{deng_arcface_2019}, transform the embeddings from Euclidean space to angular space, by removing the bias term, and normalizing both the embeddings \(x_i\) and rows of the weight matrix \( W\) within the classification layer, such that the logit is:

\begin{equation}
    W_j^T \cdot x_i = ||W_j^T|| \cdot ||x_i|| \cos(\theta_{i,j}) = \cos(\theta_{i,j})
\end{equation}

\noindent Here, \(\theta_{i,j}\) represents the angle between the embedding \(x_i\) and the \textit{j}-th column of the weight matrix \(W \in \mathbb{R}^{C \times D}\), which corresponds to the class center of the \textit{j}-th class. \(C\) denoting the number of classes, and \(D\) the embedding dimension. Additionally, an angular margin penalty \textit{m} is added to the target (ground truth) angle \(\theta_{i, y_i}\) and the logits are scaled by a scaling parameter \textit{s}. The ArcFace loss function is formulated as follows

\begin{equation}
    \mathcal{L}_{ArcFace} = - \frac{1}{N} \sum_{i=1}^{N} log \frac{e^{s \cdot \cos(\theta_{i,y_i} + m)}}{e^{s\cdot \cos(\theta_{i,y_i} + m)} + \sum_{j=1,y_i\neq j}^{C} e^{s\cdot \cos(\theta_{i,j})}} \;,
\end{equation}

\noindent where \(N\) represents the batch size.

Further approaches considered in this study build upon the ArcFace \cite{deng_arcface_2019} concept and address certain limitations. These include Sub-Center ArcFace \cite{deng_sub-center_2020}, which extends the weight matrix \(W \in \mathbb{R}^{C \times K \times D}\) by the third dimension \(K\), representing the number of sub-centers. This enforces the intra-class constraints by allowing samples to approximate proximity to one of the designated class sub-centers, which is beneficial for noisy and high intra-class variable data. ArcFace with dynamic margin adjusts the margin value according to the class sample size of the training data through a continuous mapping function (see \nameref{sec:dyn-m-af}). Li-ArcFace \cite{li_airface_2019} replaces the cosine function with a linear function, resulting in a monotonically decreasing target logit curve from 0 to \(\pi + m\). This linear approach imposes a penalty proportional to the angle between the image embedding and the class center. AdaCos \cite{zhang_adacos_2019} eliminates the need for explicit margin and scaling parameter specification, by dynamically adjusting these hyperparameters based on the number of classes in the training data. CurricularFace \cite{huang_curricularface_2020} introduces a dynamic curriculum learning strategy that initially focuses on easy samples to facilitate convergence, and gradually shifts attention to harder samples as training progresses. The difficulty of samples is determined by the angles between the image embedding and both the ground-truth and non-ground-truth class centers. Based on the difficulty and training stage, the impact of challenging negative cosine similarities is amplified through a modulation function. AdaFace \cite{kim_adaface_2022} incorporates the image quality into the margin-based metric learning loss, thereby emphasizing samples based on their image quality. This approach adjusts the margin value based on the norm of the image feature, which represents the quality of the image. As a result, hard samples with high image quality are given priority, while the impact of low-quality samples is reduced.

%
%

\section{Experimental Results \& Discussion} \label{sec:experiments}

The experiments aimed to identify the optimal combination of visual-semantic foundation model and margin-based metric learning loss for learning discriminative image embeddings enabling universal instance-level image retrieval. 
Therefore, we conducted three principle experiments: (1) a zero-shot evaluation of various image encoders (Section \ref{sec:zs-eval}), (2) an assessment of the most robust image encoders (Section \ref{sec:fm-lin-prob}), and (3) an examination of the efficacy of several metric learning losses (Section \ref{sec:loss-lin-prob}), both in linear probing the embedding model.

The linear probing experiments employed a 10-epoch training schedule using the Adaptive moment estimation (Adam) optimizer with an initial learning rate of 1e-2 and a weight decay of 1e-4. A one-epoch linear warm-up was implemented, followed by a cosine annealing scheduler with a minimum learning rate of 1e-3. \textit{M4D-35k} was used for training, making the experiments feasible by overcoming the otherwise prohibitive time and resource requirements. Input images were preprocessed by resizing the smaller edge to the target resolution of the image encoder, followed by a center crop. Metric learning losses were set with a margin of 0.5 and a scaling parameter of 30.0. The highest \(mMP@5\) across the 10 epochs was used as the primary metric to guide the decision-making process.

\subsection{Zero-shot Evaluation} \label{sec:zs-eval}

The primary objective of the zero-shot evaluation was to selectively identify robust image encoders for subsequent linear probing. This streamlined the process by excluding less effective encoders. The encoders, detailed in Table \ref{tab:zs-results}, stem from a range of foundation models, with sizes up to ViT-H for ViT's \cite{dosovitskiy_image_2020} and comparable dimensions for others. The embeddings were compressed into a 64-dimensional space by either a randomly initialized linear layer or average pooling. For SAM \cite{kirillov_segment_2023} encoders, embeddings were extracted from various network levels (see \nameref{sec:sam}), with average pooling of the ViT patch embeddings, prior to the downscaling, was found to be the most effective. 

\begin{table}[t]
\caption{Zero-shot results on the GUIEC \cite{araujo_google_2022} validation set, obtained with different foundation models. Unless stated otherwise, the encoders are employed with an image resolution of 224px. All image encoders were evaluated with two different dimensional reduction methods, random initialized linear layer or average pooling.}
\label{tab:zs-results}
\centering
\begin{tabular}{l|l|c|>{\centering\arraybackslash}p{1.5cm}>{\centering\arraybackslash}p{1.5cm}}
\hline
\multirow{2}{*}{\textbf{Method}}                        & \multirow{2}{*}{\textbf{Image encoder}} & \textbf{Pre-training}          & \multicolumn{2}{c}{\textbf{\textit{mAP}@5}} \\
                                                        &                                   & \textbf{dataset}               & pooling          & linear      \\ \hline
CLIP \cite{radford_learning_2021}                       & ViT-L/14@336px                    & WIT400M \cite{radford_learning_2021}      & 0.431            & 0.426           \\ \hline
\multirow{6}{*}{OpenCLIP \cite{ilharco_openclip_2021}}  & ViT-L/14                          & DataComp-1B \cite{gadre_datacomp_2023}    & 0.526            & 0.506           \\ \cline{2-5}
                                                        & ViT-B/14                          & \multirow{5}{*}{LAION-2B \cite{schuhmann_laion-5b_2022}}      & 0.468            & 0.454           \\
                                                        & ViT-H/14                          &                                & 0.498            & 0.509           \\
                                                        & ConvNeXt-B@256px                  &                                & 0.480            & 0.476           \\
                                                        & ConvNeXt-L@320px                  &                                & \textbf{0.584}   & 0.565           \\
                                                        & ConvNeXt-XXL@256px                &                                & 0.561            & \textbf{0.572}  \\ \hline

\multirow{2}{*}{CLIPA \cite{li_inverse_2023}}           & ViT-L/14@336px                    & \multirow{2}{*}{DataComp-1B \cite{gadre_datacomp_2023}}   & 0.583            & 0.586           \\
                                                        & ViT-H/14@336px                    &                                & \textbf{0.597}   & \textbf{0.589}  \\ \hline
\multirow{3}{*}{EVA-CLIP \cite{sun_eva-clip_2023}}      & ViT-B/16                          & \multirow{3}{*}{Merged-2B \cite{sun_eva-clip_2023}}     & 0.454            & 0.452           \\
                                                        & ViT-L/14                          &                                & 0.530            & 0.519           \\
                                                        & ViT-L/14@336px                    &                                & \textbf{0.549}   & \textbf{0.543}  \\ \hline
\multirow{3}{*}{MetaCLIP \cite{xu_demystifying_2023}}   & ViT-B/16                          & \multirow{3}{*}{MetaCLIP-2.5B \cite{xu_demystifying_2023}} & \textbf{0.422}  & 0.407           \\
                                                        & ViT-L/14                          &                                & 0.420   & \textbf{0.417}  \\
                                                        & ViT-H/14                          &                                & 0.392            & 0.392           \\ \hline
\multirow{3}{*}{SigLIP \cite{zhai_sigmoid_2023}}        & ViT-B/16@512px                    & \multirow{3}{*}{WebLI \cite{chen_pali_2023}}         & 0.535            & 0.536           \\
                                                        & ViT-L/16@384px                    &                                & 0.548            & 0.548           \\
                                                        & SoViT-400m/14@384px               &                                & \textbf{0.579}   & \textbf{0.573}  \\ \hline
\multirow{3}{*}{DINOv2 \cite{oquab_dinov2_2023}}        & ViT-B/14                          & \multirow{3}{*}{LVD-142M \cite{oquab_dinov2_2023}}      & 0.380            & 0.376           \\
                                                        & ViT-B/14@518px                    &                                & \textbf{0.436}   & \textbf{0.435}  \\
                                                        & ViT-L/14                          &                                & 0.410            & 0.396           \\ \hline
\multirow{3}{*}{SAM \cite{kirillov_segment_2023}}       & ViT-B/16                          & \multirow{3}{*}{SA-1B \cite{kirillov_segment_2023}}         & 0.111            & \textbf{0.116}  \\
                                                        & ViT-L/16                          &                                & 0.103            & 0.111           \\
                                                        & ViT-H/16                          &                                & \textbf{0.117}   & 0.113           \\ \hline
\end{tabular}%
\end{table}

Table \ref{tab:zs-results} presents the zero-shot results on the GUIEC \cite{araujo_google_2022} validation set. The results were dependent on the foundation model used, with larger encoders generally yielding better results. Notably, pre-training on DataComp-1B \cite{gadre_datacomp_2023} provided an advantage, as shown by the smaller OpenCLIP \cite{ilharco_openclip_2021} ViT-L outperforming the larger ViT-H, pre-trained on LAION-2B \cite{schuhmann_laion-5b_2022}. Convolutional Neural Networks (CNN), specifically ConvNeXt-L and -XXL, demonstrated superior performance over ViT architectures, despite being pre-trained on identical datasets. The EVA-CLIP \cite{sun_eva-clip_2023} encoders showed that increasing the input image resolution from 224px to 336px could improve the performance for the same encoder sizes. SigLIP's \cite{zhai_sigmoid_2023} SoViT-400m encoder ranked second in performance, while their smaller ViT-B outperformed all comparably sized encoders. Among all encoders, the CLIPA \cite{li_inverse_2023} ViT-H encoder achieved the highest \(mMP@5\) of 0.597.

In contrast, MetaCLIP \cite{xu_demystifying_2023}, DINOv2 \cite{oquab_dinov2_2023}, and SAM \cite{kirillov_segment_2023} encoders were not as effective as other approaches. Notably, MetaCLIP encoders underperformed relative to the original CLIP \cite{radford_learning_2021}, and larger encoders did not necessarily yield better results. The suboptimal result of DINOv2 may be attributed to the lower input image resolutions used. The DINOv2 encoders were pre-trained on images with a resolution of 518 pixels. However, owing to computational limitations, this high resolution was only feasible for the smaller ViT-B encoder, which exhibited the best performance, albeit slightly below other approaches of similar size. The weak performance of the SAM encoders can be attributed to the pixel-level pre-training methodology, which focuses on fine-grained image understanding, a strength in object detection and segmentation, but may lack global semantic understanding at the same level.

\subsection{Linear Probing - Foundation Models} \label{sec:fm-lin-prob}

\begin{table}[t]
\caption{Linear probing results on the GUIEC \cite{araujo_google_2022} validation set obtained using different image encoders as the backbone for the embedding model. The models were trained on the \textit{M4D-35k} training set using the ArcFace \cite{deng_arcface_2019} loss.}
\label{tab:fm-results}
\centering
\begin{tabular}{l|l|c|c}
\hline
\textbf{Method}                                        & \textbf{Image encoder}    & \textbf{Resolution} & \textbf{\textit{mMP}@5} \\ \hline
\multirow{3}{*}{OpenCLIP \cite{ilharco_openclip_2021}} & ViT-L/14            & 224px               & 0.660          \\
                                                       & ConvNext-L          & 320px               & 0.682          \\
                                                       & ConvNext-XXL        & 256px               & 0.700          \\ \hline
CLIPA \cite{li_inverse_2023}                           & ViT-H/14            & 336px               & 0.707          \\ \hline
EVA-CLIP \cite{sun_eva-clip_2023}                      & ViT-L/14            & 336px               & 0.672          \\ \hline
SigLIP \cite{zhai_sigmoid_2023}                        & SoViT-400m/14       & 384px               & \textbf{0.717}          \\ \hline
\end{tabular}
\end{table}

Table \ref{tab:fm-results} shows the linear probing results on the GUIEC \cite{araujo_google_2022} validation set, using the most robust image encoders from the zero-shot evaluation as the backbone for the image embedding model. The CNN architecture performed exceptionally well, outperforming both OpenCLIP \cite{ilharco_openclip_2021} and EVA-CLIP \cite{sun_eva-clip_2023} ViT \cite{dosovitskiy_image_2020} encoders, achieving a \(mMP@5\) of 0.700 for ConvNeXt-XXL. Despite the CLIPA \cite{li_inverse_2023} ViT-H encoder's leading performance in the zero-shot assessment, it was surpassed by the SigLIP \cite{zhai_sigmoid_2023} SoViT-400m, which recorded the highest \(mMP@5\) of 0.717. The SigLIP model not only outperformed the CLIPA model, but also featured a more lightweight architecture, with 400M versus 632M model parameters, enhancing the efficiency of resource utilization during training.

\subsection{Linear Probing - Metric Learning Losses} \label{sec:loss-lin-prob}

Table \ref{tab:loss-results} presents the linear probing results on the GUIEC \cite{araujo_google_2022} validation set, using the SigLIP \cite{zhai_sigmoid_2023} SoViT-400m image encoder as the backbone and a variety of margin-based metric learning losses as the loss function. The AdaCos \cite{zhang_adacos_2019} and AdaFace \cite{kim_adaface_2022} approaches did not achieve optimal results, failing to exceed the ArcFace \cite{deng_arcface_2019} benchmark, reaching an \(mMP@5\) of 0.714. In contrast, all other evaluated loss functions outperformed ArcFace, with CurricularFace \cite{huang_curricularface_2020}, and Sub-Center ArcFace \cite{deng_sub-center_2020} attaining the highest \(mMP@5\) of 0.722 and 0.720. ArcFace with dynamic margin, and Li-ArcFace \cite{li_airface_2019} yielded commendable results, reaching an \(mMP@5\) of 0.719.

The results yield the following insights: AdaFace's \cite{kim_adaface_2022} suboptimal performance may be attributed to its tendency to overfit on challenging samples (as increasingly present in GLDv2 \cite{weyand_google_2020}), as it emphasizes difficult samples of high-quality images during training. The weak results of AdaCos \cite{zhang_adacos_2019} may be caused by its hyperparameter-free nature. Since the hyperparameters (margin and scaling parameter) were optimized within the GUIEC \cite{araujo_google_2022} and used in the curation of the \textit{M4D-35k} training set, AdaCos did not provide any additional benefits. In contrast, approaches that address sample difficulty, such as CurricularFace \cite{huang_curricularface_2020} and Sub-Center ArcFace \cite{deng_sub-center_2020}, proved advantageous for the high intra-class variable \textit{M4D-35k} training set. While CurricularFace aims to learn from easier samples in the early stages and gradually introduce more challenging ones, Sub-Center ArcFace pulls easy samples towards the primary center, while hard samples are directed to non-dominant centers. This helps to mitigate intra-class constraints and increase model robustness. The use of a linear target logit curve (Li-ArcFace \cite{li_airface_2019}) or dynamic margin values that reflect the class distribution of the training set did not result in greater effectiveness than that of ArcFace \cite{deng_arcface_2019}.

\begin{table}[t]
\caption{Linear probing results on the GUIEC \cite{araujo_google_2022} validation and test set, obtained with different margin-based metric learning loss functions employed. The image embedding model used the SigLIP \cite{zhai_sigmoid_2023} SoViT-400m as backbone and \textit{M4D-35k} for training.}
\label{tab:loss-results}
\centering
\begin{tabular}{l|c|c}
\hline
\multirow{2}{*}{\textbf{Loss}}                  & \multicolumn{2}{c}{\textbf{\textit{mMP}@5}} \\ 
                                                & Val. set    & Test set \\ \hline
ArcFace \cite{deng_arcface_2019}                & 0.717       & -         \\
Sub-Center ArcFace \cite{deng_sub-center_2020}  & 0.720       & \textbf{0.721}     \\
Li-ArcFace \cite{li_airface_2019}               & 0.719       & -          \\
AdaCos \cite{zhang_adacos_2019}                 & 0.714       & -          \\
CurricularFace \cite{huang_curricularface_2020} & \textbf{0.722} & 0.715          \\
AdaFace \cite{kim_adaface_2022}                 & 0.714       & -          \\ \hline
ArcFace with dyn. margin                        & 0.719      & -           \\ \hline
\end{tabular}
\end{table}

\subsection{Evaluation on GUIEC Test Set}
In accordance with the challenge protocol, the two leading model configurations were evaluated on the GUIEC \cite{araujo_google_2022} test set to determine the final score. This involved using the SigLIP \cite{zhai_sigmoid_2023} SoViT-400m as the backbone, with linear probing of the image embedding model utilizing either CurricularFace \cite{huang_curricularface_2020} or Sub-Center ArcFace \cite{deng_sub-center_2020}. Contrary to the results on the GUIEC validation set, the configuration using Sub-Center ArcFace yielded superior performance on the test set, achieving a \(mMP@5\) of 0.721, as shown in Table \ref{tab:loss-results}.

\subsection{Comparison with SOTA Approaches}

A comparison of the performance and model size with SOTA approaches from the GUIEC \cite{araujo_google_2022} is shown in Table \ref{tab:final-results}. Leveraging the SigLIP \cite{zhai_sigmoid_2023} SoViT-400m image encoder as the backbone and solely fine-tuning the attached projection head on \textit{M4D-35k} using Sub-Center ArcFace \cite{deng_sub-center_2020} resulted in a \(mMP@5\) of 0.721 on the GUIEC test set. Notably, our approach, while employing a smaller model (based on the number of model parameters) and without end-to-end fine-tuning, trailed the GUIEC leaderboard by only 0.7 percentage points. Further, it outperformed the highest-ranked method with similar computational requirements (5th \cite{ota_5th_2022} place), achieving a substantial 3.3 percentage point improvement. In terms of deployed model size, it optimizes the total model parameters during inference by 32\% compared to the leanest approach (5th place) and reduces the number of trainable parameters by 289 times compared to the fine-tuning approaches (1st \cite{shao_1st_2022} and 2nd \cite{huang_2nd_2022} place). This achievement reflects a performance close to SOTA, surpassing the 2nd place and securing a close position behind the 1st place.

\begin{table}[t]
\caption{Performance and model size comparison of different utilized training methods (end-to-end fine-tuning or linear probing) on GUIEC \cite{araujo_google_2022} test set.}
\label{tab:final-results}
\centering
\resizebox{\textwidth}{!}{%
\begin{tabular}{l|l|c|c|c}
\hline
\textbf{GUIEC rank} & \textbf{Method} & \textbf{\# total params} & \textbf{\# train params} & \textbf{\textit{mMP}@5} \\ \hline
1st \cite{shao_1st_2022}                   & Fine-tuning           & 661M             & 661M                  & 0.728          \\
2nd \cite{huang_2nd_2022}                   & Fine-tuning           & 667M            & 667M                  & 0.709          \\
5th \cite{ota_5th_2022}                   & Linear probing     & 633M          & 1.1M                    & 0.688          \\
10th \cite{koo_10th_2022}                 & Linear probing     & 1,045M            & 22.0M                   & 0.676          \\ \hline
Own approach               & Linear probing     & 431M              & 2.3M                    & 0.721          \\ \hline
\end{tabular}
}
\end{table}

%
%

\section{Conclusion \& Future Direction}

We proposed a resource-efficient training framework for universal image embedding models capable of extracting discriminative embeddings for image retrieval at the instance-level. We have demonstrated a close to SOTA result on the GUIEC \cite{araujo_google_2022} test set while using significantly less computational resources for training. Efficiency was realized through the strategic curation of the \textit{M4D-35k} training set, the adoption of a lightweight model architecture with reduced parameter count (SoViT-400m), the application of robust pre-trained weights (SigLIP \cite{zhai_sigmoid_2023}), and the exclusive fine-tuning of the model’s projection head.

Achieving close to SOTA performance was mainly influenced by selecting the visual-semantic foundational model. The choice of an optimal margin-based metric learning loss had only a minor impact. This may be attributed to the careful selection of the training set. With \textit{M4D-35k} being optimized and adjusted, guided by a specific embedding model and training configuration, there was only limited opportunity for substantial further improvements.

Further research can be directed towards the novel large-scale multi-domain UnED \cite{ypsilantis_towards_2023} dataset. Evaluating the proposed image embedding model against the UnED benchmark would be of interest. Additionally, using the \textit{M4D-35k} training set to train the UnED baseline model would enable an evaluation of \textit{M4D-35k}'s suitability in a different setting. Alternatively, efforts could be made to surpass the UnED baseline by employing a comparably sized embedding model and a resource-efficient training methodology.

%
%
%
%

\bibliographystyle{splncs04}
\bibliography{references-010}

%
%

\newpage
\appendix

\include{appendix-010}

\end{document}

%% file: appendix-010.tex

\def\SubNumber{010}

\def\GCPRTrack{Young Researcher's Forum}

\title{Efficient and Discriminative Image Feature Extraction for Universal Image Retrieval Supplementary materials}

\ifreview
\else
	\titlerunning{Supplementary materials}

	\author{Morris Florek\inst{1}\orcidID{0009-0008-8425-5161} \and
    David Tschirschwitz\inst{1}\orcidID{0000-0001-5344-4172} \and
    Björn Barz\inst{2}\orcidID{0000-0003-1019-9538} \and
    Volker Rodehorst\inst{1}\orcidID{0000-0002-4815-0118}}
    \authorrunning{M. Florek et al.}
	
	\institute{Bauhaus-University Weimar, 99423 Weimar, Germany \email{morris.benedikt.florek@uni-weimar.de} \and 
    Carl Zeiss AG, 07745 Jena, Germany}
    
\fi

\maketitle

\section*{Appendix 1} \label{sec:data-cur}

This section presents the evaluation results on the GUIEC \cite{araujo_google_2022} validation set, achieved through linear probing of the image embedding models across various dataset configurations. The insights derived from this analysis guided the data curation process and the inclusion of datasets in the \textit{M4D-35k} training set. 

\subsubsection{GLDv2 Reduction}

\begin{table}[!b]
\caption{Linear probing results on the GUIEC \cite{araujo_google_2022} validation set. Showing different training set configurations consisting of Products-10k \cite{bai_products-10k_2020} and GLDv2 \cite{weyand_google_2020}. An \textit{x} indicates the inclusion of the dataset in the configuration. GLDv2 is used in different configurations regarding the total number of classes and maximum number of samples per class.}
\label{tab:gldv2-sampling}
\centering
\begin{tabular}{c|ccc|c}
\hline
\textbf{Products-10k \cite{bai_products-10k_2020}} & \textbf{GLDv2 \cite{weyand_google_2020}} & \textbf{\begin{tabular}[c]{@{}c@{}}max. samples \\ per class\end{tabular}} & \textbf{\# classes} & \textbf{\textit{mMP}@5} \\ \hline
x                     & -                           & -                                                                          & -                   & \textbf{0.630}      \\ \hline
x                     & x                           & 100                                                                        & 81k                 & 0.612               \\
x                     & x                           & 75                                                                         & 81k                 & 0.613               \\
x                     & x                           & 50                                                                         & 81k                 & 0.620               \\
x                     & x                           & 40                                                                         & 81k                 & 0.612               \\
x                     & x                           & 30                                                                         & 81k                 & 0.611               \\
x                     & x                           & 20                                                                         & 81k                 & 0.619               \\
x                     & x                           & 10                                                                         & 81k                 & \textbf{0.629}      \\ \hline
x                     & x                           & 10                                                                         & 38k                 & 0.641               \\
x                     & x                           & 10                                                                         & 20k                 & \textbf{0.644}       \\
x                     & x                           & 10                                                                         & 10k                 & 0.643      \\ \hline
\end{tabular}%
\end{table}

Table \ref{tab:gldv2-sampling} presents the linear probing results on the GUIEC \cite{araujo_google_2022} validation set, employing varying class counts and maximum samples per class from the GLDv2 \cite{weyand_google_2020} dataset. For benchmarking purposes, results from exclusive training on the Products-10k \cite{bai_products-10k_2020} dataset are also provided. A reduction in the maximum samples per class enhanced performance, yet it did not surpass the scores obtained from training solely on the Products-10k dataset. The optimal performance was achieved with a cap of 10 samples per class, which was maintained to guarantee a sufficient number of samples per class for the effective training of discriminative image embeddings.

Following the random reduction of the total number of utilized classes, the score improved further and reached a \(mMP@5\) of 0.644 for 20k classes. However, in the final GLDv2 \cite{weyand_google_2020} subset incorporated in \textit{M4D-35k}, only 10k classes were included for two reasons: (1) The performance disparities were minimal despite halving the training data volume. (2) A configuration of 10k classes preserved a domain distribution akin to that of the GUIEC \cite{araujo_google_2022} evaluation set.

\subsubsection*{Top-5 Dataset Combinations}

\begin{table}[t]
\caption{Linear probing results on the GUIEC \cite{araujo_google_2022} validation set, obtained with all feasible dataset combinations. The datasets added to Products-10k \cite{bai_products-10k_2020} and the subset of GLDv2 \cite{weyand_google_2020} are marked with an \textit{x}.}
\label{tab:data-combi}
\centering
\begin{tabular}{ccc|c}
\hline
\multicolumn{3}{c|}{\textbf{Added datasets}}                                 & \multirow{2}{*}{\textbf{\textit{mMP}@5}}           \\ \cline{1-3}
\multicolumn{1}{c|}{DeepFashion \cite{liu_deepfashion_2016}} & \multicolumn{1}{c|}{MET Artwork \cite{ypsilantis_met_2022}} & Shopee \cite{howard_shopee_2021} \\ \hline
\multicolumn{1}{c|}{x}           & \multicolumn{1}{c|}{}            &        & \textbf{0.652}      \\
\multicolumn{1}{c|}{}            & \multicolumn{1}{c|}{x}           &        & 0.647               \\
\multicolumn{1}{c|}{}            & \multicolumn{1}{c|}{}            & x      & \textbf{0.652}      \\ \hline
\multicolumn{1}{c|}{x}           & \multicolumn{1}{c|}{x}           &        & 0.649              \\
\multicolumn{1}{c|}{x}           & \multicolumn{1}{c|}{}            & x      & 0.650               \\
\multicolumn{1}{c|}{}            & \multicolumn{1}{c|}{x}           & x      & 0.647               \\
\multicolumn{1}{c|}{x}           & \multicolumn{1}{c|}{x}           & x      & 0.649               \\ \hline
\end{tabular}
\end{table}

Table \ref{tab:data-combi} shows the linear probing results on the GUIEC \cite{araujo_google_2022} validation set, employing all viable dataset combinations of DeepFashion \cite{liu_deepfashion_2016}, MET Artwork \cite{ypsilantis_met_2022}, and Shopee \cite{howard_shopee_2021}, with Products-10k \cite{bai_products-10k_2020} and a GLDv2 \cite{weyand_google_2020} subset being fixed. Each combination demonstrated superior performance compared to the previous \textit{M4D-35k} dataset configuration in terms of model performance. However, configurations incorporating the MET Artwork dataset exhibited the least impressive performance. The individual inclusion of DeepFashion and Shopee achieved the highest \(mMP@5\) of 0.652, surpassing even the performance of combined dataset utilization.

\subsubsection*{Swap Domain Datasets}

\begin{table}[t]
\caption{Linear probing results on the GUIEC \cite{araujo_google_2022} validation set, obtained by replacing the DeepFashion \cite{liu_deepfashion_2016} and Shopee \cite{howard_shopee_2021} datasets with datasets from the same domain. The dataset configurations consisted of Products-10k \cite{bai_products-10k_2020}, a subset of GLDv2 \cite{weyand_google_2020}, and the individual replacement dataset.}
\label{tab:data-replacement}
\centering
\begin{tabular}{l|l|c}
\hline
\textbf{Replaced dataset}                                & \textbf{Replacement}                                  & \textbf{\textit{mMP}@5} \\ \hline
\multirow{3}{*}{DeepFashion \cite{liu_deepfashion_2016}} & H\&M Personalized Fashion \cite{garcia_ling_hm_2022}  & 0.646          \\ \cline{2-3} 
                                                         & Fashion-2000k \cite{han_automatic_2017}               & 0.648          \\ \cline{2-3}
                                                         & DeepFashion2 \cite{ge_deepfashion2_2019}              & 0.647          \\ \hline
\multirow{2}{*}{Shopee \cite{howard_shopee_2021}}        & RP2k \cite{peng_rp2k_2021}                            & 0.647          \\ \cline{2-3} 
                                                         & Stanford Online Products \cite{song_deep_2016}        & 0.640          \\ \hline
\end{tabular}
\end{table}

Table \ref{tab:data-replacement} illustrates the linear probing results on the GUIEC \cite{araujo_google_2022} validation set following the substitution of DeepFashion \cite{liu_deepfashion_2016} and Shopee \cite{howard_shopee_2021} with other datasets from the same domains. Unfortunately, these alternative datasets failed to enhance performance and were thus excluded from subsequent consideration. Noteworthy is the decision to proceed with a single \textit{M4D-35k} dataset configuration, comprising Products-10k \cite{bai_products-10k_2020}, a subset of GLDv2 \cite{weyand_google_2020}, and DeepFashion. Despite the comparable performance with the individual inclusion of Shopee, this decision facilitated a broader domain variety within the \textit{M4D-35k} training set.

\subsubsection*{Add Unrepresented Domain Datasets}

Table \ref{tab:data-expansion} shows the linear probing results on the GUIEC \cite{araujo_google_2022} validation set following the incorporation of datasets from previously unrepresented domains. Recognizing the significance of the furniture \& home decor and storefronts domains within the GUIEC evaluation dataset, the Furniture-180\footnote{\url{https://www.kaggle.com/datasets/andreybeyn/qudata-gembed-furniture-180} \label{fn:f180} } and Storefronts-146\footnote{\url{https://www.kaggle.com/datasets/kerrit/storefront-146} \label{fn:s146}} datasets were integrated, despite being absence in the initial dataset list. The inclusion of datasets from the dishes, furniture \& home decor, and storefronts did not yield performance enhancements. Only the integration of the Stanford Cars \cite{krause_3d_2013} dataset led to an improvement in model performance. In its refined version, with enhanced class granularity, a \(mMP@5\) of 0.654 was achieved. 

\begin{table}[htp]
\caption{Linear probing results on the GUIEC \cite{araujo_google_2022} validation set, obtained by adding datasets from unrepresented domains to the latest M4D-35k dataset configuration (consisting of Products-10k \cite{bai_products-10k_2020}, the subset of GLDv2 \cite{weyand_google_2020} and DeepFashion \cite{liu_deepfashion_2016}). These additional datasets have been added individually.}
\label{tab:data-expansion}
\centering
\begin{tabular}{l|l|c}
\hline
\textbf{Domain}                 & \textbf{Added dataset}                                & \textbf{\textit{mMP}@5} \\ \hline
\multirow{2}{*}{Dishes}         & Food Recognition 2022 \cite{alcrowd_aicrowd_2022}     & 0.649               \\
                                & Food101 \cite{bossard_food-101_2014}                  & 0.650               \\ \hline
\multirow{2}{*}{Cars}           & Stanford Cars \cite{krause_3d_2013}                   & 0.653               \\
                                & Stanford Cars (refined)                               & \textbf{0.654}      \\ \hline
Furniture \& home decor         & Furniture-180\footref{fn:f180}                        & 0.646               \\ \hline
Storefronts                     & Storefronts-146\footref{fn:s146}                      & 0.652     \\ \hline
\end{tabular}
\end{table}

\section*{Appendix 2} \label{sec:dyn-m-af}

\begin{figure}[htp]
    \centering
    \includegraphics[width=0.65\linewidth]{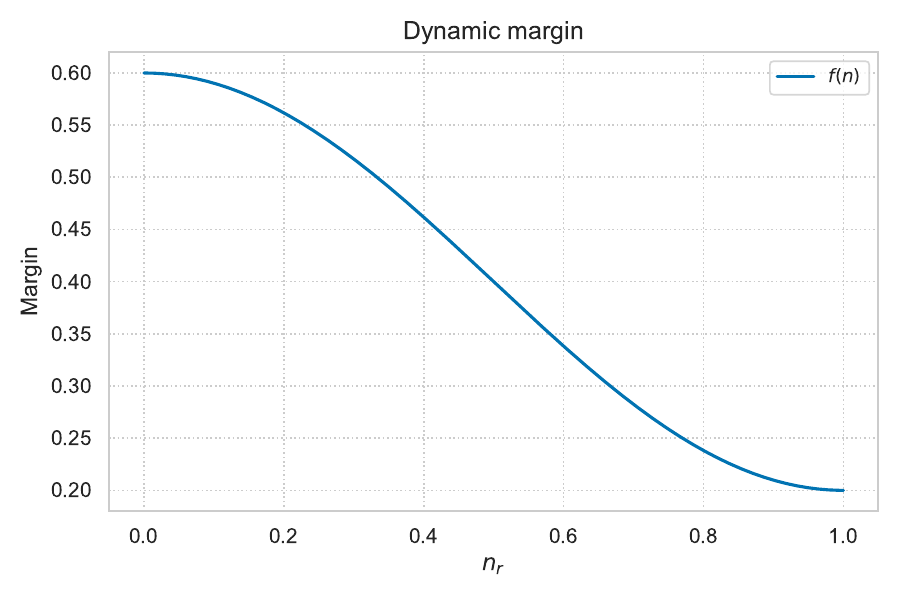}
    \vspace{-0.5cm}
    \caption{Margin mapping function \(f(n)\) with \(m_{\text{max}} = 0.6\) and \(m_{\text{min}} = 0.2\)}
    \label{fig:dyn-margin}
\end{figure}

The ArcFace \cite{deng_arcface_2019} loss distributes class centers uniformly on a hypersphere owing to the fixed margin, which may be less representative for highly unbalanced training sets. Ha et al. \cite{ha_google_2020} proposed a dynamic margin that adjusts according to class sample size, allocating larger margins to smaller, more challenging classes through a continuous function correlating class size to margin level. Inspired by this, we introduce a mapping function \(f(n)\), which correlates class size to a margin value, following a cosine curve depicted in Figure \ref{fig:dyn-margin}. The mapping function is defined as: 

\begin{equation}
    f(n) = m_{\text{min}} + 0.5 \cdot (m_{\text{max}} - m_{\text{min}}) \cdot (1 + cos(\pi \times n_r))
\end{equation}

\noindent Here, \(m_{\text{max}}\) and \(m_{\text{min}}\) are the upper and lower bounds of the margin values, while \(n_r\) denotes the rescaled class size normalized to a range between 0 and 1, defined as \(n_r = \frac{n - n_{\text{min}}}{n_{\text{max}} - n_{\text{min}}}\).

\section*{Appendix 3} \label{sec:sam}

\begin{figure}[htp]
    \centering
    \includegraphics[width=0.75\linewidth]{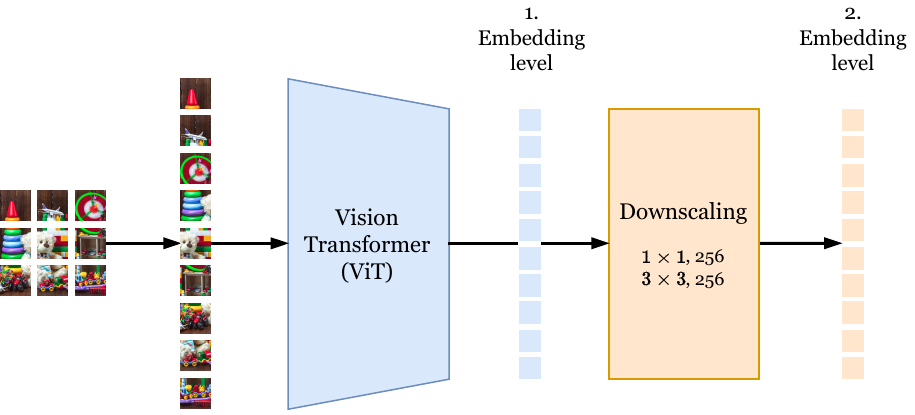}
    \vspace{-0.2cm}
    \caption{Overview of the SAM \cite{kirillov_segment_2023} image encoder and the layers from which the embeddings were extracted.}
    \label{fig:sam}
\end{figure}

\noindent Originally pre-trained for segmentation, the SAM \cite{kirillov_segment_2023} image encoder, depicted in Figure \ref{fig:sam}, encodes solely image patches without incorporating a class token. Therefore, we extracted image embeddings at two different levels within the network: before and after the downscaling of the embeddings. At the first embedding level, patch embeddings were aggregated using average pooling. At the second embedding level, patch embeddings were also aggregated using average pooling along with flattening.

\section*{Appendix 4}

Table \ref{tab:settings} outlines the final model architecture and hyperparameters utilized for linear probing in order to achieve optimal results on the GUIEC \cite{araujo_google_2022} test set. 

\begin{table}[htp]
\caption{Final model architecture and linear probing settings to obtain optimal results on the GUIEC \cite{araujo_google_2022} test set.}
\label{tab:settings}
\centering
\begin{tabular}{l|l}
Backbone                & SigLIP \cite{zhai_sigmoid_2023} SoViT-400m/14       \\
Pre-trained             & WebLI \cite{chen_pali_2023} for 45B seen samples \\
Head                    & Projection layer           \\
Output dimension        & 64                         \\
Dropout                 & 0.2                        \\
Loss                    & Sub-Center ArcFace \cite{deng_sub-center_2020}         \\
k                       & 3                          \\
m                       & 0.5                        \\
s                       & 30.0                       \\ \hline
Dataset                 & \textit{M4D-35k}                    \\
Image resolution        & \(384 \times 384\)                  \\
Transforms              & Resize, CenterCrop         \\
Batch size              & 128                        \\ \hline
Epochs                  & 10                         \\
Optimizer               & Adam                       \\
Optimizer momentum      & \(\beta_1, \beta_2 = 0.9, 0.999\)    \\
Learning rate           & 1e-2                       \\
Weight decay            & 1e-4                       \\
Learning rate scheduler & CosineAnnealing            \\
Minimum learning rate   & 1e-3                       \\
Warmup epoch            & 1                          \\
Warmup scheduler        & linear                      
\end{tabular}
\end{table}